\documentclass[runningheads]{llncs}
\usepackage[T1]{fontenc}
\usepackage{graphicx}
\usepackage{booktabs}
\usepackage[misc]{ifsym}

\usepackage{mwe}
\usepackage{float}
\usepackage{subcaption}
\usepackage{tikz,graphics,color,float,epsf}
\usepackage{enumitem}
\usepackage{amsmath}%
\usepackage{amsfonts}
\usepackage{multirow}%
\usepackage{xcolor}
\usepackage[numbers]{natbib}
\usepackage{wrapfig}

\begin{document}

\title{HADT: A Heterogeneous Multi-Agent Differential Transformer for Autonomous Earth Observation Satellite Cluster}

\titlerunning{HADT for Autonomous Earth Observation Satellite Cluster}
% If the full title of your paper is short enough to also fit in the running head, you can omit the abbreviated paper title here. You can check as follows: if you comment out the \titlerunning line, something will appear in the header of all odd-numbered pages of your PDF from page 3 onward. This something is either the full title (in which case all is well), or the error message "Title Suppressed Due to Excessive Length". If this error message appears, you're going to want to provide an abbreviated title within the \titlerunning command, because if you won't do it, Springer will do it for you.

%N.B.: Author information (both in the \author{} and \authorrunning{} command) should only be present in the Camera-Ready Version of your paper. The version that you initially submit for review, ought to be double-blind. So, when initially submitting your paper, use:
%\author{Author information scrubbed for double-blind reviewing}
% \author{Andr\'e Lauren Benjamin\inst{1} \and
% Calvin Cordozar Broadus Jr.\inst{2,3} \corr \and
% Antwan Andr\'e Patton\inst{1}\orcidID{0000-1111-2222-3333}}
\author{Mohamad A. Hady\inst{1} \and Muhammad Anwar Masum\inst{1} \and Siyi Hu\inst{2} \and Mahardhika Pratama\inst{1} \and {Zehong Cao}\inst{1} \and Ryszard Kowalczyk\inst{1,3}.}
% You may leave out the orcidID information, if you want to.
% Use \corr to indicate the corresponding author. Note the spacing around the \corr command. Only one author can be the corresponding author.

%N.B.: comment out the \authorrunning{} command for the double-blind version of your paper submitted for review. Later, if your paper is accepted, use the command for the Camera-Ready Version.
\authorrunning{M. A. Hady, et al.}
% First names are abbreviated in the running head.
% If there is one author, write 'A.L. Benjamin'.
% If there are two authors, write 'A.L. Benjamin and C.C. Broadus Jr.'
% If there are more than two authors, '[...] et al.' is used.

\institute{School of Computer Science and Information Technology, Adelaide University, Adelaide, 5095, SA, Australia. \email{\{mohamad.hady; muhammadanwar.masum; dhika.pratama; jimmy.cao; ryszard.kowalczyk\}@adelaide.edu.au}
\and
School of Electrical Engineering, Computing and Mathematical Sciences (EECMS),
Curtin University, Kent St, Bentley, 6102, WA, Australia. \email{siyi.hu@curtin.edu.au}
\and
Systems Research Institute, Polish Academy of Sciences, Warsaw, Poland.}

\maketitle              % typeset the header of the contribution

\begin{abstract}
This work addresses the problem of autonomous resource management in heterogeneous satellite cluster conducting Earth Observation (EO) missions including optical and Synthetic Aperture Radar (SAR) satellites. In autonomous operation mode, satellites are equipped with intelligent capabilities enabling real-time decision-making based on the latest conditions, while requiring minimal interaction with ground operators. Traditional scheduling approaches typically rely on mathematical models to represent satellite mission and resource management. Then, this problem is solved by using optimization algorithms. However, such solutions become less effective when the underlying models are not available, over complex, and inaccurate due to dynamic changes and uncertainties inherent in the space mission environment. A promising alternative is to reformulate the problem as a sequential decision-making process and apply model-free reinforcement learning techniques to enable adaptive and real-time resource management. To this end, we propose a novel transformer-based architecture tailored for heterogeneous satellite cluster autonomous EO Mission with relational observations-actions tokenization and differential attention mechanism. Our experimental results demonstrate significant performance improvements compared to the available baselines. Moreover, the proposed architecture exhibits strong adaptability and transferability with respect to varying numbers of satellite clusters.\footnote{\textit{This work has been accepted in ECML-PKDD'26}}

\end{abstract}

\section{Introduction}
\begin{figure}[t]
    \centering
    \includegraphics[width=0.8\linewidth]{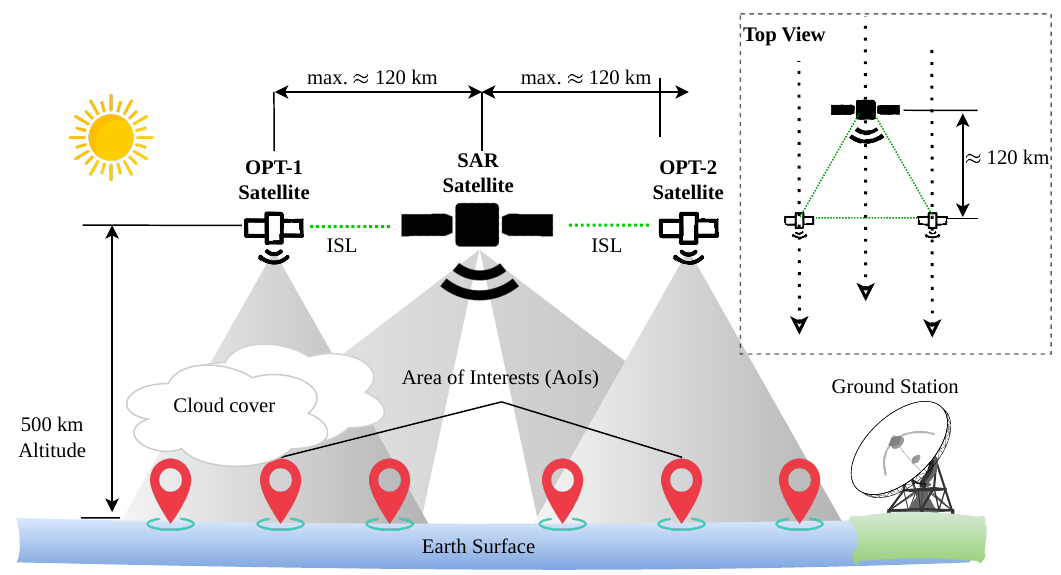}
    \caption{Illustration of A Heterogeneous Satellite Cluster for EO Mission under different cloud condition. The satellites are deployed in three different narrow orbits forming a cooperative formation. The SAR satellite is used to cover regions with higher cloud coverage that is the burden of the Optical (OPT) satellite. Managing such a mixed constellation thus requires not only efficient resource scheduling but also adaptive policies that account for diverse capabilities.}
    \label{fig:cluster}
\end{figure}

Coordinating and managing multiple Low Earth Orbit (LEO) satellites autonomously remains a challenge due to the dynamic, uncertain, and resource-constrained nature of satellite mission \cite{wang2020agile, chen2019mixedILP, stephenson2023optimal, pan2023dense}. Unlike pre-planned mission operations, autonomous EO missions require each satellite to make real-time decisions under dynamic conditions, and resource limitations, while maintaining coordinated behaviour across the entire constellation \citep{li2024mission, yang2024objective}. These challenges stem from several interacting factors: uncertainty in observation conditions (e.g., variations in target priority and cloud coverage affecting data acquisition quality), limited onboard resources such as power, data storage, attitude control, and the intrinsic non-stationarity of multi-agent settings, where each satellite’s actions continuously alter the shared environment and other agents’ states \citep{araguz2018applying, yao2019task}. Beyond these general coordination difficulties, EO missions increasingly employ heterogeneous satellite clusters by combining different payload types such as Synthetic Aperture Radar (SAR) and optical sensors \citep{cohen2017novasar, dong2024optisar, alzubairi2024spacecraft}. The SAR sensor can operate under cloud coverage and dark condition, while the optical sensor offers high-resolution imaging under clear condition. This complementary capability substantially improves temporal coverage and observation reliability yet introduces new coordination and control complexity.  The general scenario of heterogeneous cluster used in this work is illustrated in Fig.~\ref{fig:cluster}.

Traditional optimization techniques, such as mixed-integer linear programming (MILP), have been explored for constellation scheduling and resource allocation \citep{kim2024optimal}. It is effective in static conditions, rely on predefined models and are limited in their ability to adapt to time-varying environmental and operational uncertainties. A model-free Reinforcement Learning (RL) provides a promising alternative by enabling agents to learn adaptive strategies through interaction with the environment, allowing satellites to handle imaging, energy usage, and data management under uncertainty without building its mathematical model \citep{herrmann2024single, stephenson2024bsk}. However, as EO missions scale from single satellites to cooperative cluster or constellations, the decision-making problem expands from isolated optimization to distributed coordination among multiple agents — a setting more naturally modelled by Multi-Agent Reinforcement Learning (MARL) \citep{tang2024dynamic}. The existing MARL applications in satellite operations \citep{herrmann2023reinforcement, stephenson2024reinforcement} have shown encouraging results but often rely on fully centralized training or continuous communication assumptions, which simplifies real-world settings with limited access to the other agent information during execution. The Centralized Training with Decentralized Execution (CTDE) paradigms \citep{ning2024survey, hady2025multi} achieves a practical balance by enabling satellites to learn coordinated policies during centralized training while operating independently at execution time. 

Despite these advances, current MARL frameworks largely assume homogeneous agents, where all satellites share identical dynamics and observation–action structures. Recent on-policy methods such as Multi-Agent Proximal Policy Optimization (MAPPO) have demonstrated stable performance across diverse MARL environments \citep{yu2022surprising} is suitable for homogenous agent assumption. This assumption limits applicability to realistic EO missions, where sensor diversity and operational asymmetry fundamentally change the learning problem. Recent progress in heterogeneous-agent MARL, such as Heterogeneous-Agent PPO (HAPPO) \citep{zhong2024heterogeneous}, introduces separate value estimation and policy update mechanisms to address agent heterogeneity. Yet, their performance and adaptability in physically grounded, the improvement of the policy architecture remains unexplored. Especially under difficult task, the performance of this method still can be improved.

Therefore, we propose a transformer-based algorithm to handle autonomous and heterogeneous satellite clusters EO mission (HADT). Our contributions are listed as:
\begin{itemize}
    \item A problem formulation of heterogeneous satellite cluster resources management based-on Decentralized Partially Observable Markov Decision Process (DecPOMDP). We proposed scenarios incorporating three different complexity levels, including randomness and uncertainty aspect.
    \item A new transformer-based architecture algorithm applied for heterogeneous satellite cluster. We developed a new differential multi-head attention to handle noisy inputs. In addition, the HADT is a general satellite policy model, that is adaptable to the multiple heterogeneous cluster. The code is made publicly available\footnotemark{}, with a demonstration video of our experimental scenario. \footnotetext{\url{https://anonymous.4open.science/r/ECMLPKDD-2A50}}
    \item A token-based agent observations which maps the observation entities to the action entities of each satellite.
    
\end{itemize}

The rest of this paper is structured as follows: Section II presents the preliminary of this study from the problem formulation, motivation and review of the current state of the art. Section III describes the proposed method employed to solve the problem for heterogeneous satellite cluster scenarios. Section IV discusses our experimental evaluation and results. Finally, Section V concludes the paper and discusses future directions.

\section{Preliminary}
\subsection{Autonomous Heterogeneous Multi Satellite Cluster Problem Formulation}
\label{sec:2.1.}
In this subsection, a formal model of the Autonomous Heterogeneous Multi Satellite Cluster for Earth Observation (EO) mission problem is formulated. The objective is to capture as many unique high priority target as possible during in orbits. Previous works on single-satellite EO missions have formulated the problem as a sequential decision-making task in the well-known reinforcement learning framework, specifically as a Partially Observable Markov Decision Process (POMDP) \cite{stephenson2024reinforcement,stephenson2024using}. Building upon this, we formally define the multi-satellite EO mission as a Decentralized POMDP (Dec-POMDP) model as a tuple: $\mathcal{G}=\langle\mathcal{I,S,A,O,T},r,\mathcal{Z,\gamma}\rangle$.

A cluster consists of three satellites (as illustrated in Fig. \ref{fig:cluster}) and each satellite functions as an agent forming a set of agents ($\mathcal{I}$), making a decision at discrete time step ($t$) based on the current states ($\mathcal{S}$) and agent's local observations ($\mathcal{O}$). The observation is a subset of state which the agent can continuously observe, including battery level, onboard memory storage, reaction wheel speed (angular velocity), target priority, target opportunity window, cloud coverage forecast, ground station visibility windows, eclipse intervals, and simulation time. The simulation defines a finite set of possible actions ($\mathcal{A}$): \textit{1) Charging}, which involves reorienting the satellite toward the sun to maximize solar energy absorption and recharge its battery; \textit{2) Downlinking}, where the satellite transmits the collected EO image data whenever it has access to a ground station; \textit{3) Desaturating}, which ensures that the Reaction Wheels (RWs), the primary actuators for attitude control, operate within safe speed limits. If the RW speed approaches saturation, the satellite must execute a desaturation manoeuver to maintain stable attitude control and prevent uncontrollable drift; \textit{4) Capturing the $i$-th image target}, where the satellite must orient its optical imaging sensor toward a selected target-$i$ among the available targets on Earth and store it in the on-board memory. 

The instantaneous reward $r$ integrates three mission objectives: data acquisition, resource utilization, and safe operation and defined as:
\begin{equation}
r =
\begin{cases}
q_i - \rho_t + c_i, & \text{if a target is successfully captured}, \\
-\rho_t + \delta_t, & \text{if any data is downlinked}, \\
-100, & \text{if a failure occurs}, \\
-\rho_t, & \text{if only power is consumed}, \\
0, & \text{otherwise},
\end{cases}
\label{eq:reward}
\end{equation}
where $q_i \in (0,1)$ represents the priority of the target AoI at time $t$ and promotes selecting targets with greater mission importance. 

Resource usage efficiency is encouraged through three supporting terms: \textbf{1) Battery power usage} is defined as:
$\rho_t = \alpha \, \Delta Q_t \, (1 - Q_t)$, and $\Delta Q_t = Q_{t-1} - Q_t$, where $\rho_t$ penalizes excessive energy consumption ($\Delta Q_t$) times a constant $\alpha$. \textbf{2) Maximizing data downlinking} can be achieved by giving a feedback as: $\delta_t = \beta \, \Delta D_t$ and $\Delta D_t = D_t - D_{t-1}$, which rewards successful transmission of collected data. The amount of the transferred data is calculated as $\Delta D_t$ multiply by a scalar constant ($\beta$). \textbf{3) Ensuring payload correctness} either SAR or Optical (OPT) under different cloud condition:
    \begin{equation}
    c_i =
    \begin{cases}
    -1 + \sigma, & if \ \sigma < 0.5 \ \text{and captured by SAR}, \\
    \sigma, & if \ \sigma \ge 0.5 \ \text{and captured by SAR}, \\
    1 - \sigma, & if \ \sigma < 0.5 \ \text{and captured by OPT}, \\
    -\sigma, & if \ \sigma \ge 0.5 \ \text{and captured by OPT},
    \end{cases}
    \end{equation}
where, $\sigma \in (0,1)$ is the cloud coverage ratio, to guide SAR satellite to use only in cloudy conditions and optical payloads are used in clear conditions.

If the satellite encounters a failure, a fault condition is triggered, represented as:
\begin{equation}
    Failure = (b_t < m_b \space \vee \space \text{any}(\hat{\Omega}\geq\Omega_{max})),
    \label{eq:failure}
\end{equation}
where, $m_b$ is the minimum battery level to trigger a failure. This ensures safe and continuous satellite operation.

\textbf{Satellite Resources Constraints:} A single satellite has two limited resources that are considered as \textit{constraints} in our study: battery level ($b_t \in [B_{min},B_{max}]$) and data storage capacity ($d_t \in [D_{min},D_{max}]$) at any time step ($t$). At each time step, the satellite consumes electrical power, denoted as $c_{b,i}$, and stores data, represented as $c_{d,i}$. The battery is rechargeable via a solar panel. To maximize battery charging, the satellite must adjust its attitude toward the sun, which may conflict with its target imaging orientation.
Another constraint arises from attitude control, specifically the speed of the Reaction Wheels (RWs), denoted as $\hat{\Omega} \in [-\Omega_{max},\Omega_{max}]$. These wheels serve as the primary actuators for satellite attitude adjustments along the three axes ($x,y,z$). To prevent exceeding the maximum speed threshold, the satellite must periodically desaturate the wheels.
The limited resource constraints are expressed as:
\begin{equation}
    \sum^{\infty}_{t=0}{c_{b,t}}\leq b_t, \: B_{min}\leq b_t \leq B_{max} \quad
     \text{and} \quad \sum^{\infty}_{t=0}{c_{d,t}}\leq d_t, \: D_{min}\leq d_t \leq D_{max}.
\end{equation}

These constraints are incorporated into the model as \textit{failure} (Eq. \ref{eq:failure}) triggers in the reward function, resulting in penalties or negative rewards as feedback from the environment. Some other constraints, such as the communication baud rate, have a relevant impact on the system performance. However, in this work, it is assumed as fixed in time as the transmitter specification.

\subsection{Advantage of Model-free Reinforcement Learning Solution}
In traditional EO satellite mission planning, optimization problems are commonly formulated as Mixed-Integer Linear Programming (MILP) models when the system dynamics, operational constraints, and objective functions are fully known and accurately characterized \cite{chen2019mixedILP}. Under such deterministic and well-defined conditions, MILP provides a rigorous mathematical framework capable of generating globally optimal solutions for fixed planning horizons. However, in practical on-orbit autonomous operations, model inaccuracies, environmental uncertainties, and partially observable system states may degrade the effectiveness of model-based optimization. In such cases, model-free Reinforcement Learning (RL) offers a promising alternative, as it does not require explicit knowledge of the system transition model and instead learns decision policies directly through interaction with the environment. However, as RL policy employs a model to approximate the optimal policy, the scheduling solution is not the global optimal point or it is called quasi-optimal solution.
\begin{wrapfigure}{r}{0.4\textwidth}
  \begin{center}
    \includegraphics[width=0.35\textwidth]{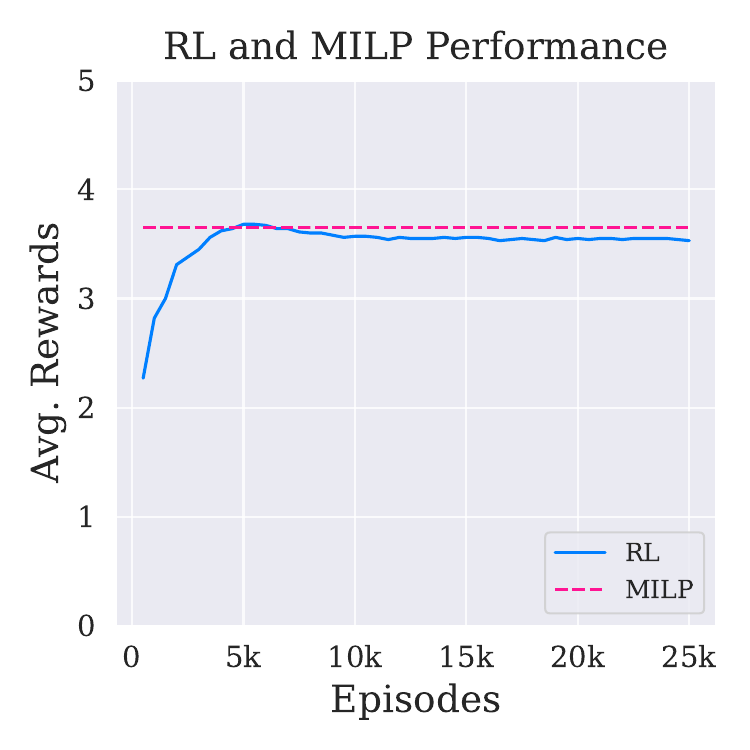}
  \end{center}
    \caption{RL and MILP Performance Comparison. RL can achieve comparable result to MILP under simple case study. Thus, model-free RL with PPO can be a potential solution to autonomous heterogeneous satellite cluster.}
    \label{fig:rl_milp}
\end{wrapfigure}

To confirm this hypotheses, we designed a simple model of satellite cluster scheduling as initial study and comparison (detailed mathematical model provided in the Supplemental Document, Section 1). The cluster is homogeneous with three optical satellites and simulated under short scheduling period (400 seconds) with only five ground targets to be captured. The constraints used in this scenario are battery, memory storage, target opportunity windows, ground station opportunity windows, eclipse condition and unique target capturing (no duplication). The resources availability is calculated for each time step (20 seconds period) and the time windows opportunity information are collected by running a Basilisk simulation, then it is fed as Proximal Policy Optimization (PPO) \cite{schulman2017PPO} RL inputs or observation. The policy outputs or action spaces are: charging, downlinking, and i-th target capturing.

From the results as shown in Fig. \ref{fig:rl_milp}, MILP can solve the problem and achieve the optimal scheduling solution with 3.65 rewards and 100\% completion rate (5/5 targets are captured). It also confirms that RL can perform comparable to MILP in a simple EO Mission case after 5000 training episodes. By adapting this preliminary study to more complex mission with stochastic dynamics, RL can be more robust to modelling errors and unforeseen operational variations, making it suitable for autonomous and dynamic EO satellite resource management.

\subsection{State of the Art of MARL Algorithm}
\begin{figure}
    \centering
    \includegraphics[width=\linewidth]{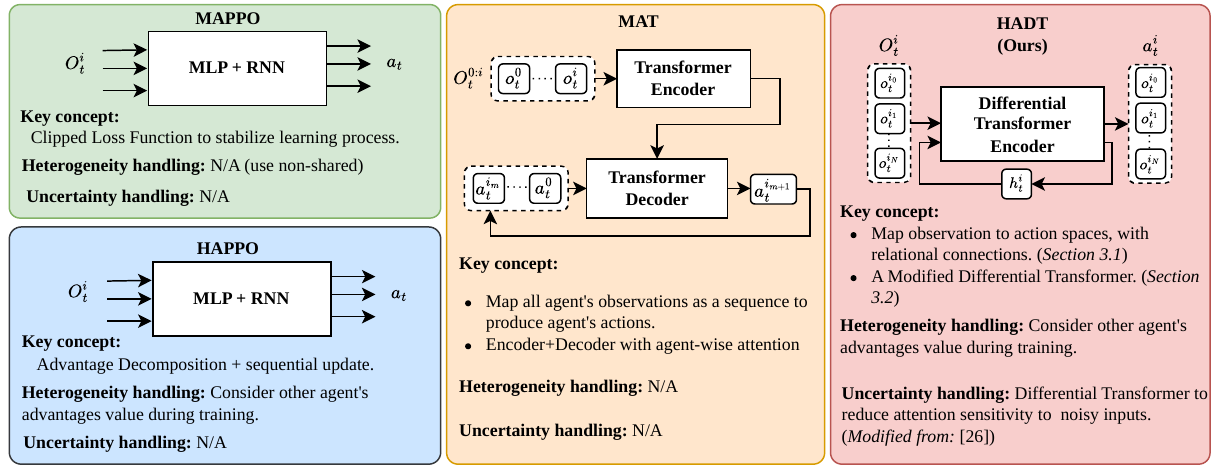}
    \caption{HADT comparison with the baseline algorithms. Our main focus is to improve the performance under uncertainty, randomness and disturbance conditions with heterogeneous agent learning feature.}
    \label{fig:HADT_baseline}
\end{figure}

To solve POMDP and Dec-POMDP of Autonomous EO Mission, a model-free approach is selected due to its flexibility to directly learn the policy, especially when the model of the system is complex and the explicit mathematical model is unavailable. In this work, three recent state-of-the-art on-policy MARL methods have been selected to be compared, which motivates us to proposed our methods: 

\textbf{1) Multi-agent PPO (MAPPO):} MAPPO is an extension of PPO designed specifically for multi-agent systems \cite{yu2022surprising}. It incorporates centralized critics and decentralized policies to improve performance in MARL tasks. MAPPO uses a single centralized critic shared by all agents, allowing the evaluation of the global state to stabilize learning and mitigate non-stationarity: $V_i^{\text{centralized}}(s) \approx \mathbb{E} \left[ \sum_{t=0}^\infty \gamma^t r_{i,t} \mid s_0 = s \right]$, where \( s \) represents the global state, \( \gamma \) is the discount factor, and \( r_t \) is the reward at time step \( t \). The loss function for the policy optimization in MAPPO is given by:

$L^{\text{MAPPO}}(\theta_i) = \mathbb{E}_t \left[ \min\left(r_t(\theta_i) \hat{A}_t, \text{clip}(r_t(\theta_i), 1-\epsilon, 1+\epsilon) \hat{A}_t\right) \right]$, where: \( r_t(\theta_i) = \frac{\pi_{\theta_i}(a_t \mid o_t)}{\pi_{\theta_i}^{\text{old}}(a_t \mid o_t)} \) is the probability ratio,\( \hat{A}_t \) is the global advantage function, and \( \epsilon \) is the clipping parameter.

\textbf{2) Heterogeneous Agent PPO (HAPPO):} This algorithm extends MAPPO by accounting for heterogeneous agents with distinct state-action spaces or roles and the sequential update scheme \cite{zhong2024heterogeneous}. It uses individual advantage functions and decentralized policies while maintaining centralized critics. In HAPPO, the centralized value function is agent-specific to handle heterogeneous agents: $V_i^{\text{centralized}}(s) \approx \mathbb{E} \left[ \sum_{t=0}^\infty \gamma^t r_{i,t} \mid s_0 = s \right]$, where \( i \) denotes the agent index and \( r_{i,t} \) is the reward specific to agent \( i \). And, the loss function for HAPPO is denoted by:
\begin{equation}
\label{eq:lossHAPPO}
    L_i^{\text{HAPPO}}(\theta_i) = \mathbb{E}_t \left[ \min\left(r_{i,t}(\theta_i) \hat{A}_{i,t}, \text{clip}(r_{i,t}(\theta_i), 1-\epsilon, 1+\epsilon) \hat{A}_{i,t}\right) \right],
\end{equation}
where \( r_{i,t}(\theta_i) = \frac{\pi_{\theta_i}(a_{i,t} \mid o_{i,t})}{\pi_{\theta_i}^{\text{old}}(a_{i,t} \mid o_{i,t})} \) and \( \hat{A}_{i,t} \) are the advantage functions of agent \( i \). And it follows the multi agent decomposition lemma and it has sequential update mechanism denoted as: $\hat{A}_{i_{1:m},t}(o_t, a_{i_{1:m},t})
=\sum_{j=1}^{m} \hat{A}_{i_j}\bigl(o_t, a_{i_{1:j-1},t}, a_{i_j}\bigr)$.

\textbf{3) Multi-Agent Transformer (MAT):} This algorithm is one of the available novel architecture that reformulates cooperative multi-agent reinforcement learning (MARL) as a sequential sequence-modelling problem by mapping agents’ joint observation sequences to optimal action sequences using a Transformer encoder–decoder framework \cite{wen2022multi}. Central to MAT is the application of the multi-agent advantage decomposition theorem, which decomposes the joint team advantage $A^{\pi}(o,a)$ into a sum of individual agent advantages conditioned on preceding agents’ actions, enabling a sequential decision process with only linear complexity in the number of agents: $A^{\pi}(o,a)=\sum_{m=1}^{n}A^{\pi}_{i_{m}}\big(o,\;a_{i_{1:m-1}},\,a_{i_{m}}\big)$, where $A^{\pi}_{i_{m}}(\cdot)$ is the advantage of agent $i_{m}$’s action given earlier agents’ choices. MAT leverages the transformer's self-attention and autoregressive decoding to model inter-agent dependencies, enjoys monotonic performance improvement guarantees, and is trained online on-policy, outperforming strong baselines such as MAPPO and HAPPO across major benchmarks.

Although MAPPO, HAPPO, and MAT can operate effectively in the CTDE paradigm, they do not have strategy to handle randomness and uncertainty of the environment. As compared in Fig \ref{fig:HADT_baseline}, MAPPO was developed for a stable learning process, then improved by HAPPO to incorporates agent-specific advantage value update to better understanding other agent's policy in heterogeneous systems. MAT maps all agent's observation as a sequence to generate agent's action, thus in our study it is categorized as a centralized paradigm. Although, it has the decentralized variant, the complete encoder-decoder architecture is more complex and preferred to be compared with our approach, which relies on only encoder architecture. Our proposed method, HADT, is designed to solve uncertainty and randomness problem occurred under heterogeneous satellite cluster EO mission. It has a unique mapping of the satellite observations to actions entities then equipped with a differential multi-head attention.

\section{Proposed Method: Heterogeneous Multi-Agent Differential Transformer (HADT)}

\subsection{Observation to Token Construction}
\begin{figure}[t]
    \centering
    \includegraphics[width=\linewidth]{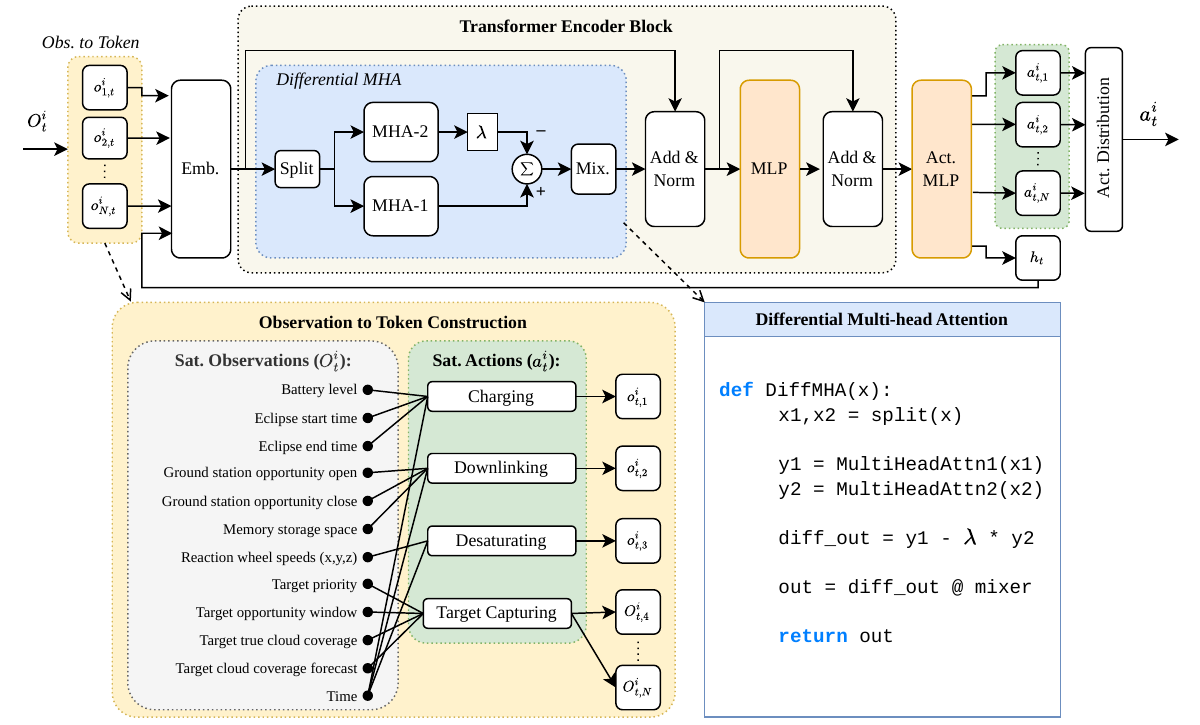}
    \caption{HADT Policy Model. It has two main components: 1) Observation to Token Construction to map observation entities to the action entities with its relation connections; 2) Differential Multi-head Attention to reduce the sensitivity of self-attention mechanism to randomness or noisy inputs.}
    \label{fig:HADTArch}
\end{figure}

Inspired by the policy decoupling mechanism in UPDeT \cite{hu2021updet}, a universal policy can be designed by constructing the transformer input token based on the number of agent's observation entities. Thus, in this work, by integrating prior knowledge of the relation between the $i-th$ satellite local observation entities ($O_t = [o_t^1, o_t^2, ..., o_t^i]$) and decision (action) entities ($a_{t}^i = [a_t^1, a_t^2, ..., a_t^i]$) at time step $t$, it can construct agent's token information by mapping the observations-actions relations as shown in Fig. \ref{fig:HADTArch}. In other words, construct the token by decoupling the observation entities into each action entities as described in Section \ref{sec:2.1.}. Since the dimensionality and semantic structure of these observations may differ across agents, a direct concatenation would be trivial for this case.

To enable structured learning, each local observation token $o_{t,N}^i$ is first projected into a shared latent space via an embedding layer $e_t^i = \text{Emb}(o_{t,N}^i)$, where $\text{Emb}(\cdot)$ denotes a learnable linear or non-linear projection. The embedding layer normalizes heterogeneous feature scales and produces fixed-dimensional token representations. Consequently, the global observation is transformed into a token set: $E_t = \{e_t^1, e_t^2, \dots, e_t^N\}$, where each token semantically represents observation-action relations. In the output side, the architecture incorporates action tokens corresponding to discrete operational modes such as charging, downlinking, target capturing, and desaturating. These action tokens allow the model to associate relational context with candidate actions and facilitate structured policy learning. The combined token representations are then processed by a Transformer encoder block composed of Differential Multi-head Attention with residual connections, layer normalization, and Multi Layer Perceptron (MLP) networks, enabling implicit inter-observations-actions and relational reasoning for each of the heterogeneous cluster satellite.

\subsection{Differential Multi-Head Attention}

To explicitly model cooperative interactions within the heterogeneous satellite cluster under uncertainty, we propose a Differential Multi-Head Attention (Diff-MHA) mechanism, inspired by \cite{ye2024differential}. The previous differential transformer architecture implements differential attention (Diff-Attn), where it has dual $Q_1,Q_2$ and $K_1,K_2$ parameters with multiple $\lambda$ parameters.  Meanwhile, our new Diff-MHA leverage a single learnable $\lambda$ parameter and introduce a new mixing parameters to achieve simpler strategy yet maintain its capability. Our proposed mechanism introduces a differential MHA operations to capture contrastive dependencies among two distinct MHA that may contain noisy informations, which can degrade the overall agent's performance.

Given the input token matrix $X \in \mathbb{R}^{N \times d}$, where $d$ is the embedding dimension, the tokens are first partitioned into two sub-representations with $X_{1,2} \in \mathbb{R}^{N \times d/2}$ and each branch is then processed by an independent Multi-head Attention (MHA) module:
    \begin{equation}
        X_1, X_2 = \text{split}(X), \quad Y_1 = \text{MHA}_1(X_1), \quad
        Y_2 = \text{MHA}_2(X_2).
    \end{equation}

This dual-attention modules capture the relational between observation's token, thus helps better representation to understand the current situation and decides the best action. For each MHA, we adopt standard MHA \cite{vaswani2017attention} with an input token matrix $X_{1,2}$, queries, keys, and values are computed as: $Q = X_{1,2}W_Q, \quad K = X_{1,2}W_K, \quad V = X_{1,2}W_V$, where $W_Q, W_K, W_V \in \mathbb{R}^{d/2 \times d_k}$ are learnable projection matrices and $d_k = \frac{d}{H}$. The scaled dot-product attention is then defined as:
    \begin{equation}
        \text{Attn}_{1,2}(Q,K,V) = \text{softmax}\left(\frac{QK^\top}{\sqrt{d_k}}\right)V.
    \end{equation}
In multi-head attention, $H$ parallel attention heads are computed and concatenated to form the outputs: 
    \begin{equation}
        Y_{1,2}=\text{MHA}_{1,2}(X) = \text{Concat}_{1,2}(\text{Attn}_{1,2}(Q,K,V)_h)_{h=1:H} W_O,
    \end{equation}
where $W_O$ is a learnable output projection matrix. Then, a differential operation is subsequently applied into both MHA's outpus:
    \begin{equation}
        Y_{\text{diff}} = Y_1 - \lambda \ . \ Y_2, \quad Y_{\text{out}} = Y_{\text{diff}} \ . \ W_{\text{mix}},
    \end{equation}
where $\lambda$ is a learnable scaling parameter controlling the influence of differential MHA. This subtraction mechanism allows the network to enhance beneficial relational signals while attenuating noisy information, thereby explicitly reduces noisy information effect. Then, the differential representation is then projected using a Mixing (Mix.) with $W_{mix}$ parameters,
where $W_{\text{mix}} \in \mathbb{R}^{d/2 \times d}$ is additional learnable projection matrix to match with the output space dimension. The output is integrated within a Transformer encoder block using residual connections and layer normalization to ensure stable training. Finally, the encoded representations are passed through an Act. (action) MLP head to produce the policy distribution $\pi(a_t \mid O_t)$, from which the action $a_t$ is sampled or selected.

The proposed Differential Multi-Head Attention mechanism enables the HADT architecture to effectively capture structured inter observations-actions dependencies, particularly it produces a suitable policy architecture for heterogeneous satellite cluster-based decision making under dynamic and constrained EO missions. This architecture eventually used as an actor encoder to approximate the policy function $\pi$ with parameters $\phi$ ($\pi_{\phi}$), where it has to be optimized following the HAPPO policy update in Eq. \ref{eq:lossHAPPO}. 

\begin{figure}[t!]
    \centering
    \begin{subfigure}[b]{0.32\textwidth}
        \centering
        \includegraphics[width=\linewidth]{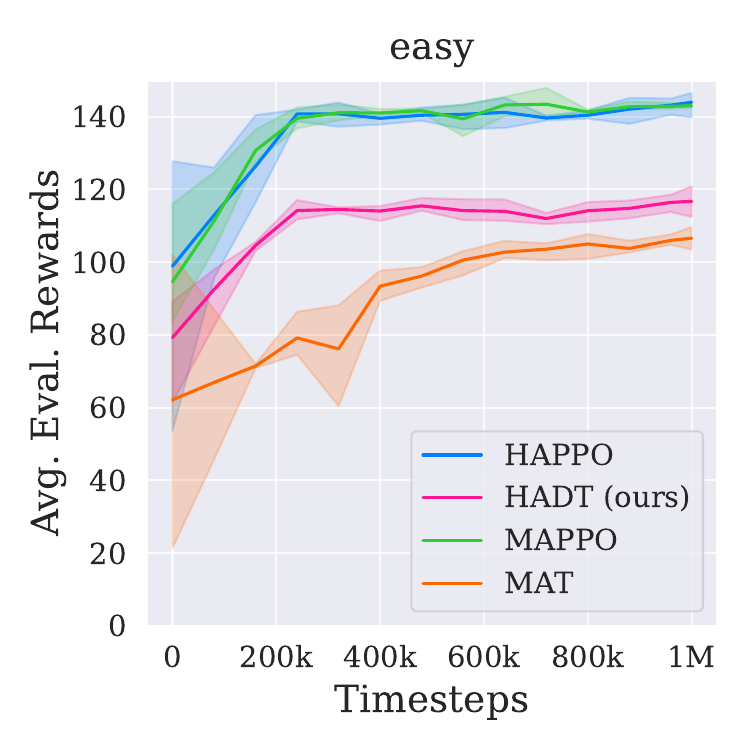}
    \end{subfigure}
    \begin{subfigure}[b]{0.32\textwidth}
        \centering
        \includegraphics[width=\linewidth]{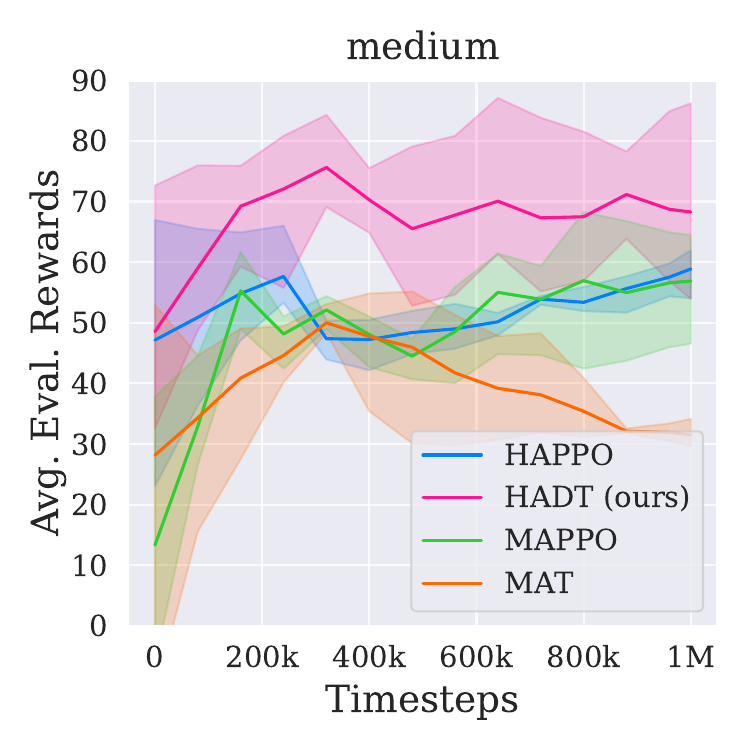}
    \end{subfigure}
    \begin{subfigure}[b]{0.32\textwidth}
        \centering
        \includegraphics[width=\linewidth]{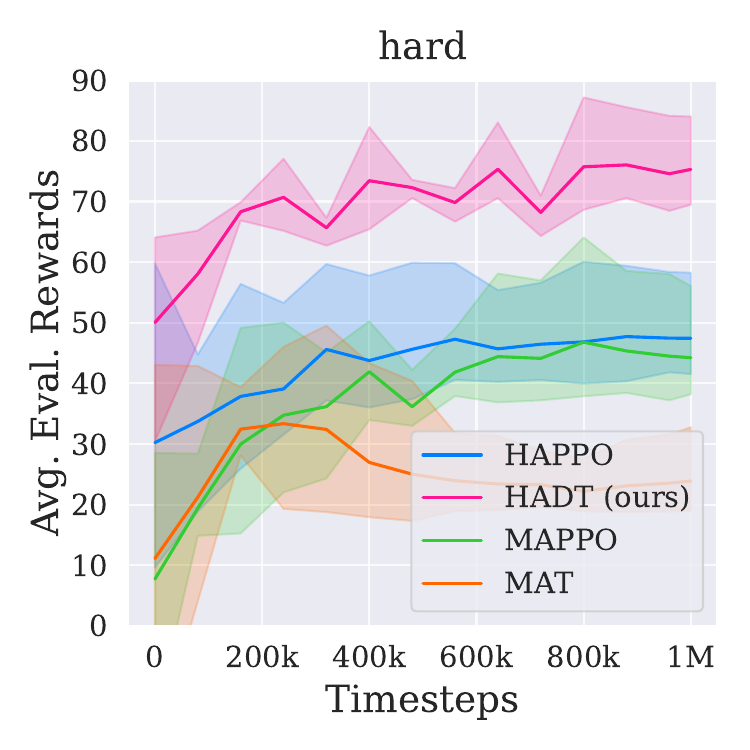}
    \end{subfigure}
    \caption{Average evaluation rewards comparison across different scenarios}
    \label{fig:avg_rewards}
\end{figure}

\section{Experimental Results}
\subsection{Scenario Description}
Our algorithm is tested under BSK-RL \cite{stephenson2024bsk} and Basilisk simulator which contain numerous simulation parameters. Due to the page limit, some of the parameters which are adjusted to match with our scenarios (e.g. orbital parameters and satellite specifications), are listed in the Supplemental Doc., Section 2. Total targets to be captured are 160 targets and the location randomized within 11 different regions. We defined three different scenarios to evaluate our proposed algorithm performance incrementally from the ideal to near-realistic simulations:

\textbf{\textit{easy}}: This scenario is designed to study the ideal satellite parameters assumption and all of resources initial conditions are set to be fully available for executing an EO mission. The battery level is fully charged (100\%), memory storage is empty (0\%), there is no randomness at the attitude disturbance and reaction wheel speed initialization). It simulates the mission with less challenge of the cluster resources availability.

\textbf{\textit{medium}}: The \textit{medium} scenario is designed to simulate the cluster under conditions of restricted resources condition. The downlink transmission speed (baud-rate) is assumed to degrade up to 30\% from its default value as in \textit{easy} case. Additionally, the battery level is directly initialized as 85\% and the memory storage is initialized as 90\%. This scenario introduces more difficulties especially to balance the memory storage utilization, if too many data collected the memory will be full and can not be properly downlinked due to the issue of the transmission speed.

\textbf{\textit{hard}}: This scenario is our main focus in this work to have a near-realistic simulation. It integrates the \textit{medium} scenario with additional randomness factors, including randomization of the initialization of battery, memory storage, disturbance and reaction wheels speed. This adds more challenge to the cluster to perform the EO mission and evaluate the generalization performance of the policy across different resources initial condition. It reflects more realistic case of the real satellite cluster EO mission, where the resources initial condition may vary depends on the states of its previous mission. The battery level is initialized randomly between 80-85\%, the memory storage is 90-100\%, disturbance is randomized with normal distribution with the scale is $10^{-4}$, and the reaction wheels are uniformly randomized between -3000 to 3000 RPM. Therefore, this scenario demonstrates more realistic challenge of the resource availability as well as its uncertainty and randomness nature at the same time.

\subsection{Performance Evaluation}
The policies of different MARL algorithms are trained under the same machine with Intel(R) Core(TM) i9-14900K CPU, 32 GB RAM, and 24 GB GPU NVIDIA GeForce RTX 4090 in this work. The algorithm hyper-parameters are listed in Supplemental Document Section 3. The training is terminated at one million training time steps with 20 parallel-processing environments (rollouts). Each rollouts has its own seed with 3 different model seeds, generating 60 unique seed combinations in our training scheme. This settings achieve at least 5000 randomized target distributions and then evaluate the performance under 500 unseen target distributions with 5 episodes and 10 rollouts each to ensure the policy's generalization performance across different target locations, priorities and cloud conditions.

\begin{figure}[t]
    \centering
    \includegraphics[width=0.8\linewidth]{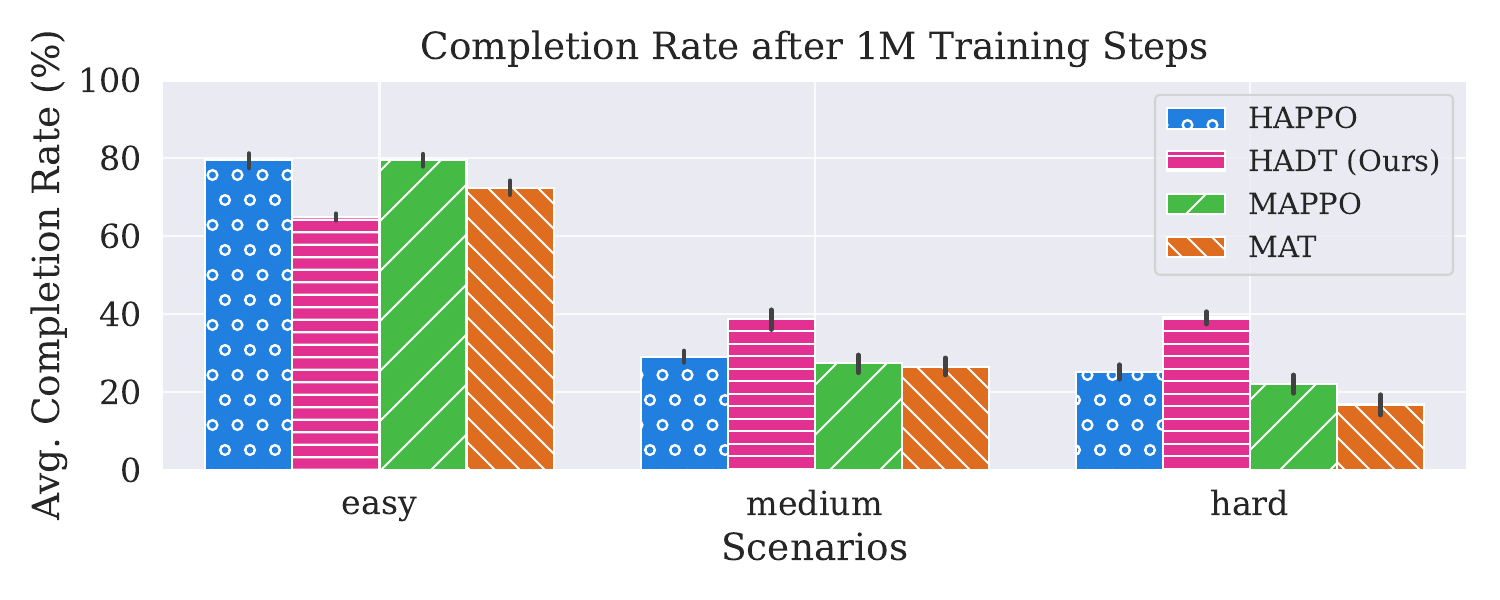}
    \caption{Average evaluation Completion Rate (CR) in different scenarios}
    \label{fig:avg_CR}
\end{figure}

The averaged total rewards metric is shown in Fig. \ref{fig:avg_rewards} and the Completion Rate (CR) is presented Fig. \ref{fig:avg_CR}. Based on those two metrics, our proposed algorithm is not suitable for \textit{easy} scenario. MAPPO and HAPPO with MLP+RNN based architecture achieve better performance in this scenario. However, this scenario is ideal case, where the resources constraints are neglected. Under more realistic scenarios, HADT outperforms the baselines under \textit{medium} and \textit{hard} scenarios. This can be achieved by its differential multi-head attention mechanism enabling more robust performance under uncertain observation with randomness. MAT is better than HADT in \textit{easy} scenarios, however, it does not have uncertainty handling mechanism in the transformer architecture, thus it can not perform well under \textit{hard} scenarios. Detailed numerical values of this experiments are presented in of Supplemental Document Section 4 (Table 4).

\subsection{The Impact of HADT}
To evaluate the significance impact of HADT, an ablation study has been performed under \textit{hard} scenario and the results are shown in Table \ref{tab:impact_study}. The results clearly demonstrate the effectiveness of HADT compared with its fundamental components and architectures. HADT achieves the highest average reward ($74.58 \pm 4.81$) and Completion Rate (CR) ($39.48 \pm 2.29$). If the differential transformer (DT) feature is deactivated it becomes the vanilla transformer variant, which attains $71.64 \pm 6.48$ in reward and $37.94 \pm 4.11$ in CR. This performance gap highlights the contribution of the differential transformer mechanism in enhancing representation learning and stabilizing coordination among heterogeneous agents. Furthermore, the margin becomes substantially larger when compared to conventional neural architectures: 1-Layer and 2-Layer MLP+RNN models with 256 nodes, whose rewards drop to $50.39$ and $47.92$, respectively, with significantly lower completion rates while increasing the number of layers. This result confirms the impact of transformer-based architecture compared with MLP+RNN. The reduced variance observed in HADT also indicates more stable learning dynamics. Overall, these results validate that the proposed algorithm enables more effective decision-making under uncertainties and dynamics, leading to superior task completion performance in complex multi-agent environments.

\begin{table}[t]
\centering
\caption{Evaluation of Different Model Architecture (CR: Completion Rate (\%); DT: Differential Transformer)}
\label{tab:impact_study}
\begin{tabular}{lcc}
\toprule
\textbf{Methods} & \textbf{Rewards (Avg. $\pm$ Std.)} & \textbf{CR (Avg. $\pm$ Std.)} \\
\midrule
HADT (ours) & $74.58 \pm 4.81$ & $39.48 \pm 2.29$ \\
Vanilla Transformer (w/o DT) & $71.64 \pm 6.48$ & $37.94 \pm 4.11$ \\
% MA + DT (w/o heterogeneous) & $... \pm ...$ & $... \pm ...$ \\
1-Layer MLP+RNN & $50.39 \pm 3.30$ & $30.17 \pm 1.74$ \\
2-Layer MLP+RNN & $47.92 \pm 2.41$ & $28.98 \pm 1.4$ \\
\bottomrule
\end{tabular}
\end{table}

\subsection{Transferability and Scalability}
\begin{figure}[t]
    \centering
        \begin{subfigure}[b]{0.43\textwidth}
        \centering
        \includegraphics[width=\linewidth]{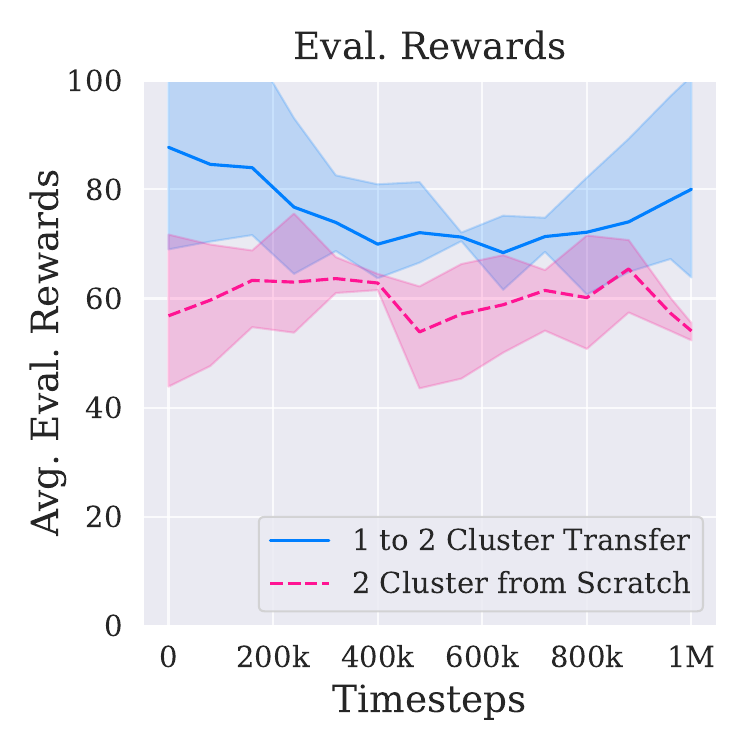}
    \end{subfigure}
    \begin{subfigure}[b]{0.43\textwidth}
        \centering
        \includegraphics[width=\linewidth]{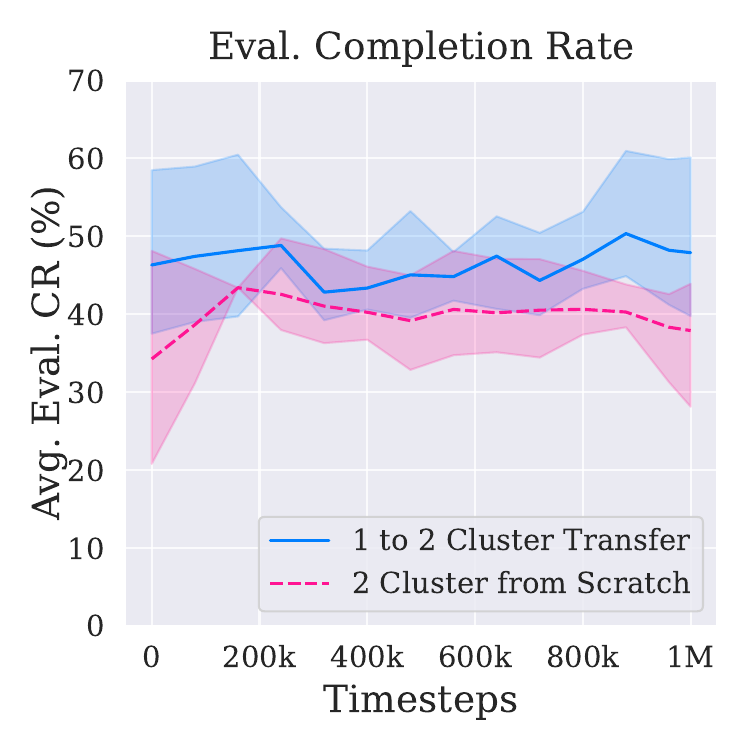}
    \end{subfigure}
    \caption{HADT Policy Transfer from 1 to 2 Cluster}
    \label{fig:transfer}
\end{figure}

The last experiment is the extension from single to two clusters. In this study, we evaluate the HADT transferability and adaptability to the new cluster configurations under \textit{hard} scenario. We add a new cluster separated in 45 degrees different orbital offset. Then, the policy is duplicated for the new cluster and then trained with 1 million time steps to adapt the policy performance to the new task. In other words, the pre-trained HADT policy for single cluster is transferred and adapted to the new  two cluster task. The results are shown in Fig. \ref{fig:transfer}. Compared with training of \textit{2 Cluster from Scratch}, the \textit{1 to 2 Cluster Transfer} where the HADT policy is duplicated for the new cluster, it can achieve better performance around 20 points rewards gap. This confirms the transferability and adaptability performance of our HADT algorithm. Also, by using two different cluster it can improve the completion rate (CR) around 10\%.

\section{Conclusion}
Transformer-based architecture has a promising future to solve problems in realistic multi-agent reinforcement learning problem. From our study, we have evaluated a new differential multi-head attention can improve the vanilla transformer to help agent learn under dynamic and uncertain condition with randomness. Moreover, the observation tokenization is a key strategy to guide transformer model by constructing the relation between observations and actions spaces. Therefore, the HADT algorithm outperforms MAPPO, HAPPO, and MAT under \textit{medium} and \textit{hard} scenarios where it is designed to mimic near-realistic conditions in EO mission. The HADT also is a potential candidate to be used for transfer learning or fine-tuning to different cluster numbers. However, if it is considered to be implemented as onboard autonomy algorithm in the future, the model contains numerous parameters that may insufficient due to the limited computational power. One of our future research can be the integration of parameter efficient fine-tuning, where it is currently one of developing research topic especially for large model in large language modelling or image processing field and it can be adopted to online multi-agent reinforcement learning research area.

\begin{credits}
\subsubsection{\ackname} 
This work has been supported by the SmartSat CRC, whose activities are funded by the Australian Government’s CRC Program. This work use an open-source realistic satellite simulator (Basilisk and BSK-RL) that is actively developed by Dr. Hanspeter Schaub and team at AVS Laboratory, University of Colorado Boulder. Also, the authors would like to express their sincere gratitude to BAE Systems for their invaluable support and collaboration throughout this research.

% \subsubsection{\discintname}
% ......................
\end{credits}
%
% ---- Bibliography ----
%
% BibTeX users should specify bibliography style 'splncs04'.
% References will then be sorted and formatted in the correct style.
%
\bibliographystyle{ieeetr}
\bibliography{references}
%% Note that this preceding line implies that you store your BibTeX references in a file called 'mybibliography.bib'. If you instead store your references in a file with a different name, for instance 'references.bib', the preceding line should read '\bibliography{references}'. Whatever you do, DO NOT put the file name extension .bib inside the \bibliography command; this will trip up LaTeX compilers. 
%
% If you do not want to use BibTeX, you can also type up the bibliography exactly as you see fit, using the following structure:
% \begin{thebibliography}{8}
% % Note that this number 8 reserves an amount of space (equal to the natural width of the given number) for the label of your references; if you have more than 9 references, you will want to change this number to 18. If you have more than 19 references, this number is best changed to 88. If you have more than 99 references, I salute you.
% \bibitem{ref_article1}
% Author, F.: Article title. Journal \textbf{2}(5), 99--110 (2016)

% \bibitem{ref_lncs1}
% Author, F., Author, S.: Title of a proceedings paper. In: Editor,
% F., Editor, S. (eds.) CONFERENCE 2016, LNCS, vol. 9999, pp. 1--13.
% Springer, Heidelberg (2016). \doi{10.10007/1234567890}

% \bibitem{ref_book1}
% Author, F., Author, S., Author, T.: Book title. 2nd edn. Publisher,
% Location (1999)

% \bibitem{ref_proc1}
% Author, A.-B.: Contribution title. In: 9th International Proceedings
% on Proceedings, pp. 1--2. Publisher, Location (2010)

% \bibitem{ref_url1}
% LNCS Homepage, \url{http://www.springer.com/lncs}, last accessed 2023/10/25
% \end{thebibliography}
\end{document}